\documentclass[conference]{IEEEtran}

\usepackage{cite}
\usepackage{amsmath,amssymb,amsfonts}
\usepackage{algorithmic}
\usepackage{graphicx}
\usepackage{textcomp}
\usepackage{xcolor}
\usepackage{url}
\def\BibTeX{{\rm B\kern-.05em{\sc i\kern-.025em b}\kern-.08em
    T\kern-.1667em\lower.7ex\hbox{E}\kern-.125emX}}

\begin{document}

\bstctlcite{IEEEtran:BSTcontrol}
\setlength{\abovedisplayskip}{3pt} 
\setlength{\belowdisplayskip}{3pt}

\title{CLaSP: Learning Concepts for Time-Series Signals from 
Natural Language Supervision}

\author{\IEEEauthorblockN{Aoi Ito}
\IEEEauthorblockA{\textit{Hosei University} \\
    Tokyo, Japan \\
    24t0002@cis.k.hosei.ac.jp}
\and
\IEEEauthorblockN{Kota Dohi}
  \IEEEauthorblockA{\textit{Hitachi Ltd.} \\
    Tokyo, Japan \\
    kota.dohi.gr@hitachi.com}
\and
\IEEEauthorblockN{Yohei Kawaguchi}
  \IEEEauthorblockA{\textit{Hitachi Ltd.} \\
    Tokyo, Japan \\
    yohei.kawaguchi.xk@hitachi.com}
}

\maketitle

\begin{abstract}
This paper presents CLaSP, a novel model for retrieving time-series signals using natural language queries that describe signal characteristics. 
The ability to search time-series signals based on descriptive queries is essential in domains such as industrial diagnostics, where data scientists often need to find signals with specific characteristics. 
However, existing methods rely on sketch-based inputs, predefined synonym dictionaries, or domain-specific manual designs, limiting their scalability and adaptability.
CLaSP addresses these challenges by employing contrastive learning to map time-series signals to natural language descriptions. 
Unlike prior approaches, it eliminates the need for predefined synonym dictionaries and leverages the rich contextual knowledge of large language models (LLMs).
Using the TRUCE and SUSHI datasets, which pair time-series signals with natural language descriptions, we demonstrate that CLaSP achieves high accuracy in retrieving a variety of time series patterns based on natural language queries. 
\end{abstract}


\section{Introduction} \label{intro}
Time-series data is ubiquitous across domains such as industrial diagnostics, healthcare, finance, and climate science.
Retrieving specific patterns in time-series data based on natural language descriptions is a fundamental yet challenging task~\cite{Roitman2013}.
For example, in industrial diagnostics, data scientists often search for signals with specific characteristics (e.g., ``the signal contains noise throughout and increases exponentially'') to develop and validate diagnostic algorithms.
This requires a system that effectively links natural language queries to time-series data for efficient retrieval.

Existing methods for time-series retrieval face significant limitations. 
Query-by-Sketch approaches~\cite{10.1145/3183713.3193547,7883519} require users to manually draw signal patterns, making them impractical for complex or ambiguous queries.
Domain-specific systems~\cite{10.1145/506443.506460,10.1145/3209978.3210069} rely on predefined signal types and patterns, which restricts their applicability to diverse datasets.
Imani et al.~\cite{10.1145/3308560.3317308} proposed a method for natural language-based retrieval depend heavily on manually designed pattern classes, quantification formulas, and synonym dictionaries.
These limitations make such approaches difficult to scale as the variety of time-series patterns and natural language expressions increases.

To address these challenges, this paper proposes CLaSP, a domain-independent method for retrieving time-series signals based on natural language descriptions. 
CLaSP leverages contrastive learning to encode the relationships between time-series signals and their corresponding natural language descriptions. 
By training on datasets such as TRUCE and SUSHI, which pair time-series signals with descriptive annotations, the model generalizes across diverse signal patterns and linguistic expressions. 
Additionally, CLaSP leverages the conceptual knowledge of large language models (LLMs), allowing it to handle diverse queries without relying on predefined synonym dictionaries. 
Experiments demonstrate that CLaSP achieves high accuracy in retrieving diverse time-series patterns from natural language queries.

\begin{figure*}[t]
    \includegraphics[width=\linewidth]{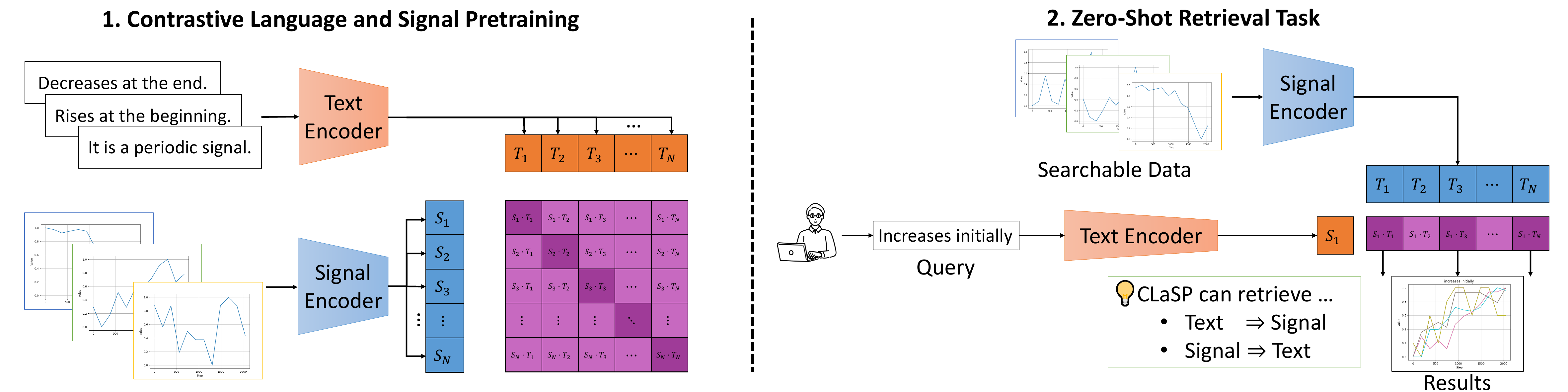}
    \caption{Overview of CLaSP.  
    The CLaSP model leverages contrastive learning to train encoders for time-series signal data and natural language expressions. By aligning their embeddings in a common feature space, the model maximizes cosine similarity for correctly paired data, as illustrated in the left half of the figure.
    In retrieval tasks, the model calculates the cosine similarity between the query embedding and embeddings of candidate time-series signals. The time-series signal with the highest similarity is returned as the result, as shown in the right half of the figure.
    }
    \label{fig:overview}
\end{figure*}

\section{Related Work} \label{relation}
Various tasks and solutions leveraging the characteristics of time-series signal data have been proposed, including signal pattern classification, forecasting, missing value interpolation, and anomaly detection~\cite{zhou2021informerefficienttransformerlong, jin2024timellmtimeseriesforecasting, liu2024unitimelanguageempoweredunifiedmodel, ansari2024chronoslearninglanguagetime, gao2024unitsunifiedmultitasktime}.
In parallel, approaches that process time-series signal data using natural language have also been explored~\cite{jhamtani-berg-kirkpatrick-2021-truth, masry2023unichartuniversalvisionlanguagepretrained, dohi}. 
TRUCE~\cite{jhamtani-berg-kirkpatrick-2021-truth} is a model that learns numerical patterns and temporal information in time-series data. 
However, it struggles with patterns or natural language expressions not included in its training data. 
UniChart~\cite{masry2023unichartuniversalvisionlanguagepretrained} extracts information expressed in natural language from graphical representations of time-series data, such as line charts. 
While effective for graphical representations, this approach is less suitable for raw sensor data, which is the primary focus of this study.
Dohi et al.~\cite{dohi} proposed a method for automatic caption generation for signal data, but they did not consider the retrieval task targeted in this study.
The most closely related work to the natural language-based time-series retrieval proposed in this paper is the method by Imani et al.~\cite{10.1145/3308560.3317308}. 
Their approach employs conventional text-based retrieval techniques for time-series data but heavily relies on predefined dictionaries containing signal patterns and their synonyms. 
This reliance significantly limits the flexibility and scalability of their solution.

Contrastive learning has recently emerged as a powerful technique for bridging data and natural language. Its success has been demonstrated in fields such as image and audio processing through models like CLIP~\cite{radford2021learningtransferablevisualmodels} and CLAP~\cite{elizalde2022claplearningaudioconcepts}. 
CLIP aligns images with captions using contrastive learning, enabling the representation of visual concepts in natural language and achieving high versatility without task-specific training.
Similarly, CLAP extends this principle to audio data by aligning audio features with natural language, enabling flexible class prediction across multiple downstream tasks.
In the domain of time-series data, contrastive learning has also been applied. 
For example, TENT~\cite{zhou2023tentconnectlanguagemodels} aligns IoT data, such as videos or LiDAR data, with descriptive words using contrastive learning.
Moon et al.~\cite{moon2022imu2clipmultimodalcontrastivelearning} proposed a method that aligns motion sensor data from inertial measurement units with natural language. 
However, these approaches are designed for specific domains or tasks, limiting their general applicability.
In this study, we aim to address the limitations of existing methods by developing a domain-independent and generalized model for time-series data retrieval using natural language representations. 
Our approach leverages contrastive learning to bridge time-series signal data and natural language effectively, enabling flexible and scalable retrieval capabilities for data scientists.

\section{Proposed Method} \label{proposed}

\subsection{Overview of the Proposed Method} \label{proposed_overview}
To describe the shapes of time-series signal data using natural language expressions, the proposed model, CLaSP (Contrastive Language and Signal Pretraining), leverages contrastive learning to establish relationships between time-series signal features and natural language expressions (Fig.~\ref{fig:overview}).  
The model takes time-series signal data and corresponding text annotations as inputs, which are processed by separate encoders for signals and text.  
The encoded features are linearly projected into a common feature space, where contrastive learning aligns pairs of time-series signals and natural language expressions within a batch.  
The pretrained encoders and projection layers can then generate embeddings for both time-series signals and natural language expressions, enabling zero-shot classification tasks based on the learned feature space.

\subsection{Acquisition of Knowledge Represented in Natural Language Using LLMs} \label{LLM}

A key aspect of the proposed approach is learning perceptual concepts about time-series signals represented in natural language through LLMs.  
By incorporating LLMs and leveraging supervised data expressed in natural language, the method eliminates the need for pre-prepared dictionaries of words and synonyms, as required in Imani et al.'s approach~\cite{10.1145/3308560.3317308}.  
As demonstrated in CLIP and CLAP, the use of LLMs enables the model to generalize beyond the specific patterns or expressions seen in input queries or training data, making it well-suited for handling a wide range of time-series signal patterns and accommodating natural language expressions that vary in style.
Given the scarcity of annotated datasets for time-series signals in natural language, LLMs are also employed as text encoders to expand the range of expressions the model can handle.

\subsection{Contrastive Learning Between Time-Series Signals and Natural Language} \label{contrastive}

To learn natural language expressions that appropriately describe the characteristics of various time-series signal patterns, contrastive learning is utilized.  
The training dataset consists of $N$ pairs of time-series signal data $X_s$ and corresponding natural language text $X_t$, denoted as $\lbrace{X_s, X_t}\rbrace_i \quad (i \in [0, N))$, where $i$ is the index of the training data.  
The pairs $X_s$ and $X_t$ are input into the time-series signal encoder $f_s(.)$ and the text encoder $f_t(.)$, respectively.  
\begin{equation}
\hat{X}_s = f_s(X_s), \quad \hat{X}_t = f_t(X_t)
\end{equation}
$\hat{X}_s \in \mathbb{R}^{N \times V}$ represents the time-series signal representation of dimension $V$, and $\hat{X}_t \in \mathbb{R}^{N \times U}$ represents the text representation of dimension $U$.

Next, learnable linear projections are applied to project the time-series signal representation $\hat{X}_s$ and the text representation $\hat{X}_t$ into a common feature space of dimension $d$.  
\begin{equation}
E_s = L_s(\hat{X}_s), \quad E_t = L_t(\hat{X}_t)
\end{equation}
$E_s \in \mathbb{R}^{N \times d}$ and $E_t \in \mathbb{R}^{N \times d}$, where $L_s$ and $L_t$ represent the linear projections for time-series signals and natural language text, respectively.

Within this common feature space, embeddings $(E_s, E_t)$ of time-series signal data and natural language expressions can be compared.  
Specifically, the similarity between time-series signal data and natural language expressions is calculated as follows:  
\begin{equation}
C = \tau \cdot (E_t \cdot E_s^\top)
\end{equation}
Here, $\tau$ is a temperature parameter used for scaling logits.  
The similarity matrix $C \in \mathbb{R}^{N \times N}$ contains $N$ correct pairs (pairs of time-series signal data and corresponding natural language expressions) on its diagonal, while the off-diagonal elements include $(N^2 - N)$ incorrect pairs.  
In the similarity matrix $C$ shown in purple in the left half of Fig.~\ref{fig:overview}, the dark purple diagonal elements represent the $N$ correct pairs.

To learn a common feature space that enables the comparison of embeddings from different modalities (time-series signals and natural language), a cross-entropy loss $\mathcal{L}$ based on the similarity matrix $C$ is defined.  
Through this loss, the time-series signal encoder and the text encoder are jointly trained along with their linear projections.  
\begin{equation}
\mathcal{L} = 0.5 \cdot (\ell_{\text{t}}(C) + \ell_{\text{s}}(C))
\end{equation}

Here, $\ell_{\text{t}}(C)$ and $\ell_{\text{s}}(C)$ represent the cross-entropy losses along the text and time-series signal axes, respectively, and are defined as follows:  
\begin{equation}
\ell_{\text{t}}(C) = -\frac{1}{N} \sum_{i=0}^{N} \log(\text{diag}(\text{softmax}(C))),
\end{equation}
\begin{equation}
\ell_{\text{s}}(C) = -\frac{1}{N} \sum_{i=0}^{N} \log(\text{diag}(\text{softmax}(C^\top))).
\end{equation}

\subsection{Zero-Shot Retrieval Task} \label{zero}

In the zero-shot retrieval task, CLaSP employs a common feature space to compute the similarity between an input query, which can be either natural language text or time-series signal data, and the target data to be retrieved.  
For a task where natural language text describing the desired time-series signal pattern is used as a query, the model first generates embeddings for both the query and the candidate time-series signal data using the pretrained encoders and projection layers. 
The cosine similarity between the query embedding and the embeddings of all candidate time-series signals is then calculated, with higher similarity indicating a closer match to the query pattern.  
The same process can be applied in reverse, where time-series signal data is used as the query to identify the most relevant natural language description.  

CLaSP's common feature space enables effective zero-shot retrieval by aligning representations of time-series signals and natural language, even for patterns or expressions not explicitly encountered during training.

\section{Experiments} \label{experiment}

\subsection{Experimental Conditions} \label{condition}

For the time-series signal encoder, we used Informer~\cite{zhou2021informerefficienttransformerlong}, and for the text encoder, we employed T5~\cite{raffel2023exploringlimitstransferlearning}.
Informer is a model designed for Long Sequence Time-series Forecasting (LSTF), based on Transformer~\cite{10.5555/3295222.3295349}. It is specifically tailored to address challenges such as quadratic time complexity and memory consumption.  
The Informer used in this experiment was trained from scratch.
T5 is a framework that unifies various natural language processing tasks, such as summarization, question answering, and text classification, into the Text-to-Text format.  
In this experiment, we used the T5-Small model available on Hugging Face~\footnote{\url{https://huggingface.co/google-t5/t5-small}}.

\begin{table}[tb]
\centering
\caption{Examples of four queries based on the class label ``sawtooth wave'' defined in the SUSHI data.}
\renewcommand{\arraystretch}{1.0} 
\setlength{\tabcolsep}{6pt}       
\begin{tabular} {p{2.5cm} | p{5.0cm}}
\hline
Sample caption for SUSHI\cite{sushi2024} & The signal is showing a periodic pattern that repeats at regular intervals, like a sawtooth wave.              \\ \hline
Captions limited to a maximum of 9 words & The signal shows a sawtooth wave. \\ \hline
``The signal is {[}Class Label{]}.'' & The signal is sawtooth wave.\\ \hline
{[}Class Label{]} & sawtooth wave \\ \hline
\end{tabular}
\label{fig:new_label}
\end{table}

\begin{table}[tb]
\centering
\caption{Example results of a time-series signal data retrieval task using CLaSP with natural language text as a query.}
\renewcommand{\arraystretch}{1.0} 
\setlength{\tabcolsep}{6pt}       
\begin{tabular} {p{3.5cm} | p{4.0cm}}
\hline
\textbf{Query} & \textbf{Results} \\
\hline
\raggedright{hits peak at the end\\(Text included in the TRUCE test data)} & 
\vspace{0.2pt}\raisebox{-0.5\height}{\includegraphics[width=4.5cm]{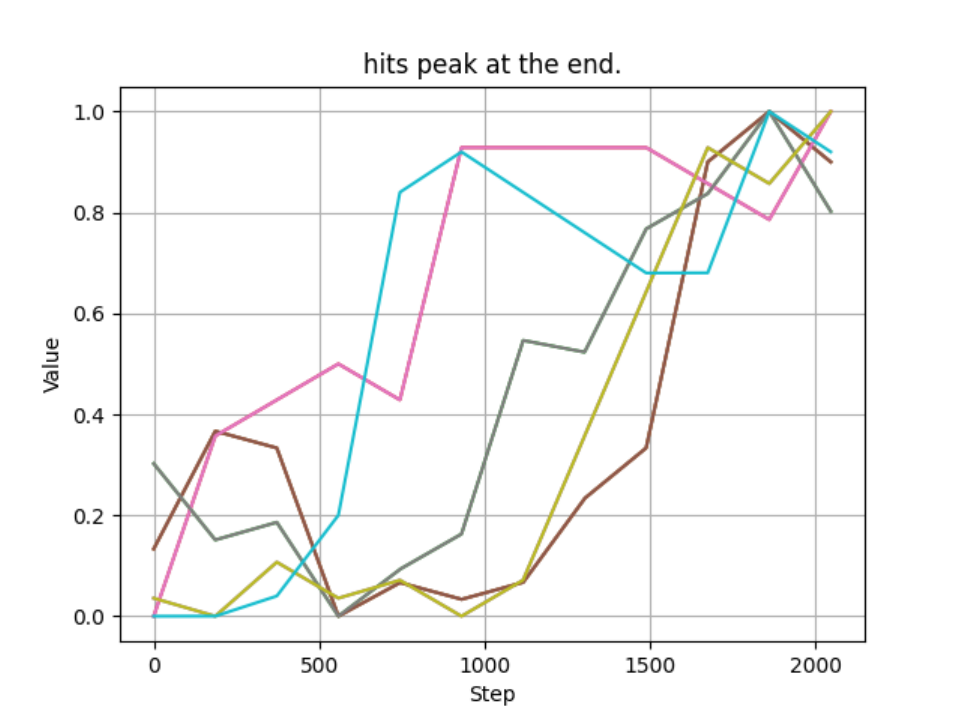}}\vspace{0.2pt} \\
\hline
\raggedright{This direction mimics the curve of a negative cubic function, starting with a decline, experiencing a rise in the middle and finally resuming the descent. This forms an S-shape rotated by 90 degrees. Besides, the signal is interrupted sporadically by large positive spikes.\\(Text contained in SUSHI test data)} & 
\vspace{0.2pt}\raisebox{-0.5\height}{\includegraphics[width=4.25cm]{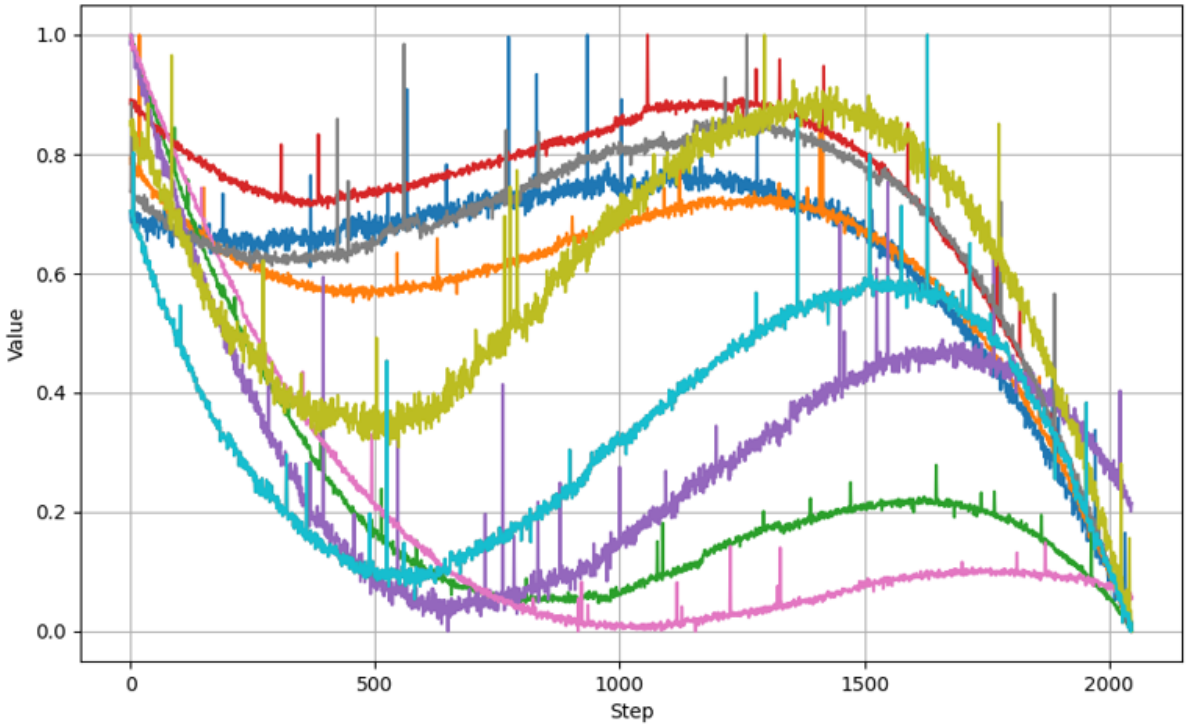}}\vspace{0.2pt} \\
\hline
\vspace{0.2pt}\raisebox{-0.5\height}{\includegraphics[width=3.5cm]{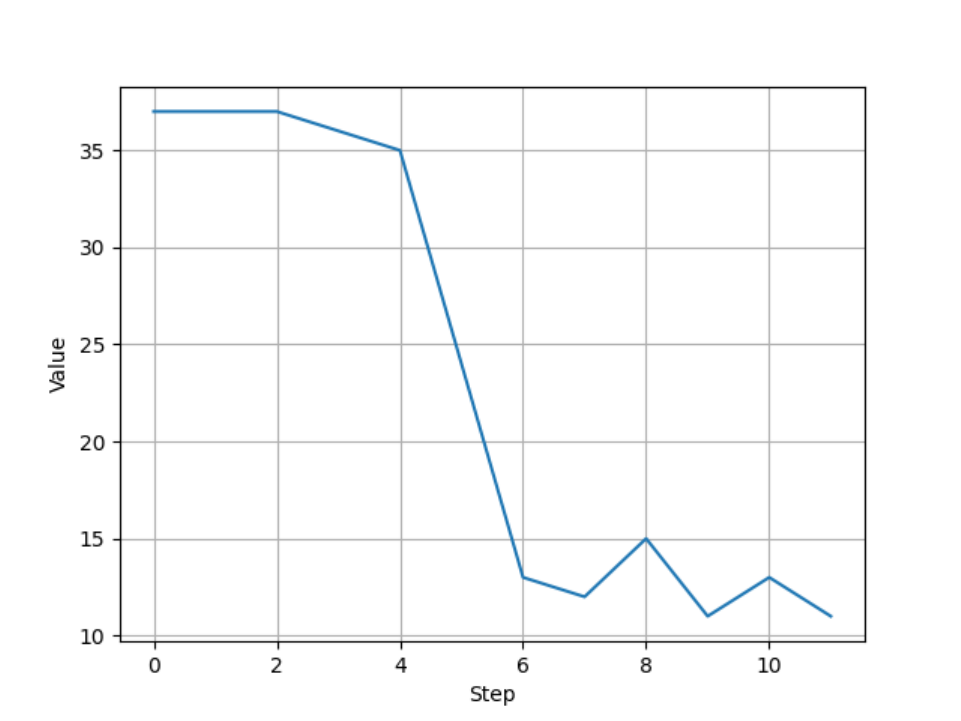}}\vspace{0.2pt} & 
\begin{itemize}
    \item gradually decreases
    \item steady decline then small uptick and down again
    \item downward trend starting quarter of way 
    \item decreases after the peak
\end{itemize} \\
\hline
\vspace{0.2pt}\raisebox{-0.5\height}{\includegraphics[width=3.5cm]{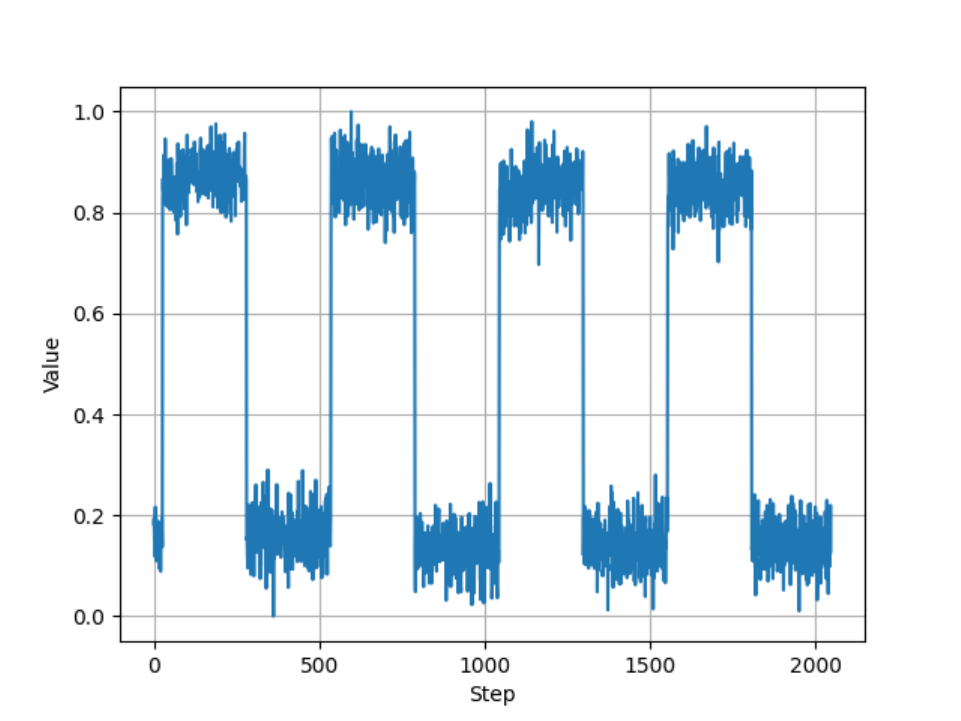}}\vspace{0.2pt} & 
\begin{itemize}
    \item Signal shows a permanently repeating sequence at regular intervals, similar to a square wave. Furthermore, the signal is covered by a large amount of noise throughout.
    \item The signal exhibits a periodic tendency to reverberate at regular intervals, comparable to square waves. Besides, the signal is sporadically interrupted by large amounts of noise.
    \item Signal shows a periodic pattern that repeats at regular intervals, similar to a square wave. Furthermore, the signal is covered by a large amount of noise throughout.
\end{itemize} \\
\hline
\end{tabular}
\label{fig:results}
\end{table}

\begin{table*}[tb]
\caption{mAP@10 for time-series signal data retrieval based on text queries with CLaSP. The columns for ``Sentence-BERT'' and ``DistilBERT'' display the mAP@10 values calculated based on objective correctness assessments. Specifically, the query text and the captions of the retrieved time-series signals were transformed into embedding vectors using Sentence-BERT or DistilBERT. Correctness was then determined based on whether the cosine similarity between these vectors exceeded the threshold, and mAP@10 was calculated. In contrast, the ``Human'' column shows the mAP@10 values calculated based on subjective correctness assessments by annotators.}
\centering
\begin{tabular}{l|ccccc}
\hline
              & \multicolumn{1}{l}{Sentence-BERT ($ts=0.5$)} & \multicolumn{1}{l}{Sentence-BERT ($ts=0.8$)} & \multicolumn{1}{l}{DistilBERT ($ts=0.5$)} & DistilBERT ($ts=0.8$) & Human\\ \hline
TRUCE         & 0.458 & 0.136 & \textbf{1.000} & 0.491 & 0.571\\
SUSHI         & \textbf{0.982}  & \textbf{0.571} & \textbf{1.000}& \textbf{0.992} & \textbf{0.848}\\
TRUCE + SUSHI & 0.954 & 0.556 & 0.959 & 0.960 & 0.842\\ 
\hline
\end{tabular}
\label{tab:result}
\end{table*}

\begin{table}[tb]
\caption{mAP@10 for each of the four queries generated based on SUSHI class labels.}
\centering
\resizebox{\columnwidth}{!}{
\begin{tabular}{l|cc}
\hline
Query & \multicolumn{1}{l}{Trend/Periodic} & \multicolumn{1}{l}{Fluctuation} \\ \hline
Sample caption for SUSHI\cite{sushi2024}       & \textbf{1.000} & \textbf{0.857}               \\
Captions limited to a maximum of 9 words       & 0.815 & \textbf{0.857}               \\
``The signal is {[}Class Label{]}.''           & 0.550 & \textbf{0.857}               \\
``{[}Class Label{]}''                          & 0.795 & 0.714               \\ \hline
\end{tabular}
}
\label{tab:result_class}
\end{table}

In this experiment, we used the datasets provided by TRUCE~\cite{jhamtani-berg-kirkpatrick-2021-truth} and SUSHI~\cite{sushi2024}.  
Both datasets include natural language annotations for each time-series signal data, describing the characteristics of the data.

TRUCE is a set of 1,900 synthetic and stock price time-series, annotated with captions in three natural languages, each limited to a maximum of nine words.
These time-series signals are publicly available and split into training, validation, and test sets in a ratio of $8:1:1$.

The SUSHI dataset is composed of time-series signal data, natural language texts describing their patterns, and corresponding class labels.  
This dataset was designed for training and evaluating foundational models for the natural language-based time-series signal retrieval tasks and caption generation tasks discussed in this paper.  
Similarly to TRUCE, the SUSHI dataset was split into training, validation, and test sets in a ratio of $8:1:1$ for this experiment.  
Unlike existing datasets, the annotation texts in SUSHI describing the patterns of time-series signals do not include context-dependent explanations.  
While the time-series signals in TRUCE consist of only 12 data points per signal, the signals in SUSHI consist of 2048 data points, which is commonly used as the length of the time window in many sensor data analyses.  

Furthermore, class labels are assigned to the time-series signals in SUSHI based on their patterns.  
The class labels are divided into three categories:  
\begin{itemize}
    \item \textbf{Trend}: Represents the overall trend of the signal (e.g., linearly increasing, exponentially decreasing).  
    \item \textbf{Periodic}: Represents periodic components of the signal (e.g., sine wave, sawtooth wave).  
    \item \textbf{Fluctuation}: Represents stochastic fluctuation components (e.g., large noise, positive spikes).
\end{itemize}

\subsection{Evaluation Method} \label{eval}

This experiment aims to evaluate the ability of CLaSP to handle zero-shot retrieval tasks. 
Specifically, we focus on retrieving time-series signal data using text queries that describe the desired patterns of the signals. 
To measure retrieval performance, we adopt Mean Average Precision at 10 (mAP@10) as the primary evaluation metric. 
mAP@10 evaluates both the relevance and ranking of the top-10 retrieved results, with higher mAP@10 values indicating more accurate retrieval performance.

For evaluation, we used the test datasets of TRUCE and SUSHI.  
In the task of retrieving time-series signal data using text as a query, the annotation texts included in the SUSHI test dataset were used as queries.  
For the top-10 search results retrieved by the model, we calculated the cosine similarity between the embeddings of the input query and the annotation texts originally attached to the retrieved results. 
If the cosine similarity exceeded a threshold ($ts$), the query and the corresponding annotation text were considered similar, indicating successful retrieval.
To ensure fairness, we used two text encoders (Sentence-BERT~\cite{reimers-2019-sentence-bert} and DistilBERT~\cite{sanh2020distilbertdistilledversionbert}) that were not employed during the training of CLaSP. 
This approach allowed us to verify whether the common feature space learned by CLaSP was robust and appropriate.
In addition to the objective evaluation, a subjective evaluation was conducted to assess the relevance of the top-10 retrieved results based on human judgment.
Three annotators independently judged the correctness of the retrieved results, and mAP@10 was calculated for each annotator. 
The average mAP@10 across all annotators was then computed to provide a comprehensive measure of retrieval performance.

Furthermore, we evaluated the performance of the zero-shot retrieval task using queries created based on the class labels defined in the SUSHI dataset.   
Four types of queries were employed:
1. Sample captions for each class describing the trend, periodicity, and fluctuations of time-series signals, as published in SUSHI~\cite{sushi2024}. 
2. Captions constrained to a maximum of 9 words, to align with the annotation restrictions in TRUCE.
3. Template sentences using class labels (e.g., ``The signal is [Class Label].'').  
4. Class labels only.  
During evaluation, the retrieved signals were compared to the class labels used in query creation to determine whether the retrieval results matched the expected outcomes.

It is important to note that CLaSP cannot reference the annotation texts attached to the time-series signal data during retrieval.
The annotation texts and labels are solely used for evaluation purposes and are not utilized in generating search results. 

\section{Results} \label{result}

This section presents the results of time-series signal data retrieval using the proposed CLaSP model.
Examples of retrieval results are shown in Table~\ref{fig:results}, demonstrating that CLaSP accurately captures the general characteristics of time-series signals, such as increase/decrease patterns and periodicity, based on query inputs.
Additionally, when the query is a time-series signal, CLaSP successfully identifies appropriate sentences from the search set that describe the signal data.

Table~\ref{tab:result} summarizes the mAP@10 results for time-series signal data retrieval tasks where natural language text was used as queries.
Each row corresponds to the type of test dataset that contains the annotation texts used as queries.  
While the features embedded by the text encoder differ, leading to some variations between the performances of Sentence-BERT and DistilBERT, the mAP@10 values for searches using texts from the SUSHI dataset all exceeded $0.5$.  
This indicates that more than half of the search results aligned with the expected outcomes.  
For TRUCE, however, the performance degraded compared to SUSHI.  
This discrepancy is likely due to differences in annotation styles: while CLaSP effectively learns the overall shape of time-series signals, nuances related to scale in TRUCE annotations were ignored, leading to reduced retrieval accuracy.

Table~\ref{tab:result_class} provides an analysis of the retrieval performance across four types of queries generated based on the SUSHI class labels.  
The results demonstrate that the majority of queries, particularly those involving stochastic fluctuations, maintain consistent performance regardless of variations in natural language expressions.
Importantly, these findings highlight the capability of the CLaSP model to deliver accurate retrieval results even for queries with expressions not encountered during training. 
This suggests that the CLaSP model eliminates the need for predefined dictionaries summarizing expressions for time-series signals, which were essential in the retrieval method proposed by Imani et al.~\cite{10.1145/3308560.3317308}.

\section{Conclusion} \label{conclusion}

In this paper, we proposed the CLaSP model, which leverages contrastive learning to retrieve time-series signal data using domain-independent natural language expressions commonly employed by data scientists.  
By establishing a common feature space between time-series signal data and natural language expressions, the proposed method enables direct comparisons between the two, supporting zero-shot retrieval tasks where either time-series signal data or natural language expressions serve as queries.


\bibliographystyle{IEEEtran}
\bibliography{refs}

\end{document}